\documentclass{article}

\PassOptionsToPackage{numbers, compress}{natbib}

\usepackage{graphicx}     
\usepackage[dvipsnames]{xcolor}
\usepackage{hyperref}       
\usepackage{url}            
\usepackage{booktabs}       
\usepackage{amsfonts}       
\usepackage{nicefrac}       
\usepackage{microtype}      

\usepackage{bm}
\usepackage{graphicx}
\usepackage{parskip}
\usepackage{algorithmic} 
 \usepackage{float}
\usepackage{algorithm}
\usepackage{mathtools}
\usepackage{amssymb}
\usepackage{wrapfig}
\usepackage{setspace}
\usepackage{placeins}
\usepackage{tikz}
\usetikzlibrary{fit,calc}
\usepackage{ulem}
\usepackage{array}
\usepackage{multirow}

\usepackage{bbm}

\usepackage{amsmath,amsfonts,bm}
\usepackage{xspace}

\newcommand{\methodName}{NegVerse\xspace}

\def\eqref#1{equation~\ref{#1}}

\def\1{\bm{1}}

\DeclareMathAlphabet{\mathsfit}{\encodingdefault}{\sfdefault}{m}{sl}
\SetMathAlphabet{\mathsfit}{bold}{\encodingdefault}{\sfdefault}{bx}{n}

\newcommand{\E}{\mathbb{E}}

\newcommand{\R}{\mathbb{R}}

\renewcommand{\thefootnote}{\fnsymbol{footnote}}

\definecolor{darkgreen}{rgb}{0.0, 0.6, 0.4} 

\newcommand{\highlight}[1]{\colorbox{green!10}{\textcolor{darkgreen}{#1}}}
\newcommand{\highlightres}[1]{\begingroup\setlength{\fboxsep}{1pt}\colorbox{CadetBlue!10}{\textcolor{Violet}{#1}}\endgroup}
\newcommand{\highlightap}[1]{\begingroup\setlength{\fboxsep}{1pt}\colorbox{lightgray!10}{\textcolor{darkgray}{#1}}\endgroup}

\newcommand{\redout}[1]{\textcolor{red}{\sout{#1}}}

\definecolor{DarkBlue}{rgb}{0.0, 0.0, 0.55}
\definecolor{lightorange}{RGB}{255, 230, 200} 
\definecolor{darkorange}{RGB}{204, 85, 0} 

    \usepackage[final]{neurips_2024}

\usepackage[utf8]{inputenc} 
\usepackage[T1]{fontenc}    
\usepackage{hyperref}       
\usepackage{url}            
\usepackage{booktabs}       
\usepackage{amsfonts}       
\usepackage{nicefrac}       
\usepackage{microtype}      
\usepackage{xcolor}         

\title{Generating Diverse Negations from Affirmative Sentences}

\author{%
  Darian Rodriguez Vasquez\textsuperscript{\dag}   \\ 
  University College London \\ 
  \texttt{darian.vasquez.23@ucl.ac.uk} \\
  \And
  Afroditi Papadaki\textsuperscript{\ddag} \\ 
  Legal \& General   \\ 
  \texttt{afroditi.papadaki@landg.com} \\
}

\begin{document}


\maketitle

\footnotetext{\textsuperscript{\dag} Research conducted as an MSc student in the Department of EEE at University College London.}
\footnotetext{\textsuperscript{\ddag} Research conducted as an academic in the Department of EEE at University College London.}
\renewcommand{\thefootnote}{\arabic{footnote}}

\begin{abstract} 
Despite the impressive performance of large language models across various tasks, they often struggle with reasoning under negated statements. Negations are important in real-world applications as they encode negative polarity in verb phrases, clauses, or other expressions. Nevertheless, they are underrepresented in current benchmarks, which mainly include basic negation forms and overlook more complex ones, resulting in insufficient data for training a language model. 
In this work, we propose \methodName, a method that tackles the lack of negation datasets by producing a diverse range of negation types from affirmative sentences, including verbal, non-verbal, and affixal forms commonly found in English text. We provide new rules for masking parts of sentences where negations are most likely to occur, based on syntactic structure and use a frozen baseline LLM and prompt tuning to generate negated sentences.  We also propose a filtering mechanism to identify negation cues and remove degenerate examples, producing a diverse range of meaningful perturbations. Our results show that \methodName outperforms existing methods and generates negations with higher lexical similarity to the original sentences, better syntactic preservation and negation diversity.  The code is available in \url{https://github.com/DarianRodriguez/NegVerse}.
\end{abstract}

\section{Introduction}\label{sec:intro}
Recent advancements in natural language processing (NLP) have enhanced various applications such as text generation \cite{wang2023automated}, translation  \cite{translation} and summarization \cite{basyal2023text}, but handling negation remains a significant challenge \cite{hossain-etal-2020-analysis}. Negations are crucial for reasoning and effective communication, as they express denial, contradiction, and absence. This is especially important in critical fields like biomedicine, where misinterpreting negated conditions can have serious consequences. For example, Large Language Models (LLMs) identifying acute bleeding \cite{bleeding} have misclassified cases with negated phrases, revealing bias and a limited understanding of negations \cite{garcíaferrero2023dataset, li2024logical}.
\vspace{-0.05in}

Despite their importance, existing literature has established that language models struggle with negated sentences in tasks such as cloze completion, NLI, QA, and classification \cite{asher2024stronghallucinationsnegationfix,hossain-etal-2020-analysis,li2024logical}. For example, the work in \cite{truong2023language} found an inverse scaling trend among models such as GPT-J, GPT-$3$, Flan-T$5$, GPT-Neo, and OPT (ranging from $125$M to $6$B parameters), where larger models tend to perform worse on negation tasks and often produce incorrect answers with high confidence. Similarly, \cite{jang-etal-2022-becel} and \cite{kassner2020negatedmisprimedprobespretrained} demonstrated that models like BERT, RoBERTa, GPT-2, BART, and T5 frequently generate identical outputs for opposite statements and misinterpret sentences, such as classifying "The man in the blue shirt is relaxing on the rocks" as entailing "A man is \highlightres{not} wearing a blue shirt".
\vspace{-0.05in}

Negations are also underrepresented in most benchmark datasets, both in terms of frequency and complexity. In particular, the works in \cite{hossain2022analysis} and \cite{hossain-etal-2020-analysis} show that general-purpose English corpora, such as reviews, conversations, Wikipedia, and books, contain between $22.6\%$ and $29.9\%$ sentences with negations. In contrast, some natural language inference benchmarks have around $8.7\%$, while other datasets, such as COPA \cite{copa} and QQP \cite{cer-etal-2017-semeval}, contain $0.8\%$ and $8.1\%$ respectively. 
\vspace{-0.05in}

To improve negation understanding in NLP models, it is crucial to expand annotated datasets to cover various types of negation across different domains \cite{Morante_Blanco_2021}. Transformer-based models, such as RoBERTa \cite{liu2019robertarobustlyoptimizedbert} and BERT \cite{devlin-etal-2019-bert}, often struggle with negations due to their underrepresentation in training data \cite{hossain2022analysis}. Current benchmarks primarily focus on verbal negations, lacking syntactic and morphological negations \cite{garcíaferrero2023dataset,hossain2022analysis}. Although some existing methods address verbal negations \cite{hossain-etal-2020-analysis} or use rule-based augmentation \cite{helwe-etal-2022-tina}, they still cover only a limited range of negation types.
\vspace{-0.05in}
\paragraph{Contributions:} To address this issue, we introduce \methodName, a method that generates a diverse range of syntactic and morphological negations, including non-verbal, verbal, and affixal forms, to enrich the training datasets. \methodName (a) keeps the produced negated data closely aligned with the original sentences by employing a masking strategy at both token and subtree levels; and (b) addresses the shortage of affixal negation datasets and other negation forms, by assembling $362$ unique sentences using LLama-$2$ and other sources, such as COPA \cite{copa} and SNLI \cite{ bowman-etal-2015-large}. We introduce an efficient masking strategy to insert negations while maintaining sentence fluency. Additionally, a new filtering mechanism is used to exclude degenerate outputs, capturing key negation cues effectively. We use a GPT-2-based model to generate negated sentences and implement a filtering mechanism that screens the generated negations for closeness, duplicates, and validity. 
We provide extensive empirical evidence of our \methodName's efficiency and improved performance using relevant criteria such as closeness, diversity, and text quality \cite{ madaan2021generatecounterfactualscontrolledcounterfactual, sachdeva-etal-2024-catfood,wu2021polyjuicegeneratingcounterfactualsexplaining} on various datasets against state-of-the-art baselines.

\section{Related Work}\label{sec:related_work}

LLMs have excelled in various tasks \cite{basyal2023text,  translation, kamalloo2023evaluating, wang2023automated}, but they consistently struggle with understanding negated sentences \cite{jang2022large}, which limits their reasoning abilities \cite{truong2023language} and sometimes worsens with the model size. 
Current solutions, such as syntactic data augmentation using Semgrex patterns \cite{hosseini2021understanding} 
 and the TINA method \cite{helwe-etal-2022-tina}, aim to enhance LLMs' robustness to negations in textual entailment tasks by augmenting training datasets with grammatically correct negated instances. However, they face errors in complex sentences. 
Other approaches like \cite{singh-etal-2023-nlms} generate negated data using tense patterns and keywords, while \cite{garcíaferrero2023dataset} uses WordNet to create true/false sentences. Nevertheless, these methods are not adaptable across diverse datasets.
Polyjuice \cite{wu2021polyjuicegeneratingcounterfactualsexplaining} generates sentence perturbations but produces nonsensical outputs and handles a limited range of negation types.

The work in \cite{hossain2022leveraging} transforms negated sentences into affirmative ones using sentence pairs and back-translation yet it falls short compared to human understanding. 
Similar to the aforementioned approaches, our goal is to produce new negated sentences to augment the existing datasets. However, in contrast to these methods, our proposed approach generates a wider range of negations -- including verbal, non-verbal, and affixal forms -- from affirmative sentences. It employs an efficient masking strategy to maintain fluency and structural preservation, resulting in outputs that align lexically better with the original sentences and overcome the limitations of earlier methods.

\section{Problem Formulation}\label{sec:prob_form}

We consider a dataset $\mathcal{D} = \{ (x_i, \mathbf{c}_i, \hat{\mathcal{X}}_i) \}_{i=1}^m$, where $x_i$ denotes an affirmative sentence, $\mathbf{c}_i = \{ c_i^{(j)} \}_{j=1}^n$ is the corresponding context vector, and $\hat{\mathcal{X}}_i = \{ \hat{x}^{(1)}, \hat{x}^{(2)}, \dots,\hat{x}^{(n)} \}$ the set of all the valid ground-truth negated sentences. The affirmative sentences lack any negation and do not include information guiding the construction of its negation. Each context $\mathbf{c}_i$ includes $n$--structured prompts with placeholders denoted as \textsf{\color{blue}[BLANK]}, indicating where the negation should be applied within a sentence $x_i$. The set $\hat{\mathcal{X}}_i$ contains the respective valid negated sentences corresponding to context $\mathbf{c}_i$.

 \vspace{-0.06in}
 Our goal is to learn a language generator model $g \in \mathcal{G}$, parametrized by a vector $\bm \theta \in \Theta$, that, given an affirmative sentence $x$ and context $\bm {c}$, produces a set of negated versions $\mathcal{\hat{X}}_{\textsf{gen}}$ that closely approximates the ground truth negated set $\hat{\mathcal{X}}$. This is equivalent to solving 
\vspace{-0.05in}
\begin{equation}\label{eq:initial_objective}
    \min_{\bm\theta \in  \Theta} \mathop{\E}\limits_{\mathcal{D}}\bigg[ 
    \frac{1}{m}\sum_{\hat{x} \in \hat{\mathcal{X}}}\ell(\hat{x}_{{\textsf{gen}}},\hat{x})
    \bigg],  
    \end{equation}     
where $\ell: \Delta^{m-1}\times\Delta^{m-1}\to \R_+$ is the loss function with $\Delta$ representing the probability simplex and $\hat{x}_{\textsf{gen}}$ being the output of the generator model given a pair of an affirmative sentence and a context vector, formally defined as $\hat{x}_{\textsf{gen}}=g(\bm\theta;x,c)$, with $x \in \mathcal{X}$. 
 \vspace{-0.03in}

One challenge with the objective in Eq. \ref{eq:initial_objective} is that the generated set $\mathcal{\hat{X}}_{\textsf{gen}}$ may contain incoherent or irrelevant sentences, leading to nonsensical outputs that reduce model effectiveness. Additionally, the generator requires a context vector to determine the appropriate negation placement, which might not be available for every input. 
To address these issues, we next propose masking spans in sentences at positions where negation is appropriate, which are used to generate structured prompts that approximate the missing context $\mathbf{c}$, thereby enabling the model to produce accurate and contextually appropriate negations, even without the original context. We also provide a filter that selects only the contextually accurate and meaningful negations $\mathcal{\hat{X}}_{\textsf{f}}$ from $\mathcal{\hat{X}}_{\textsf{gen}}$, i.e., $\mathcal{\hat{X}}_{\textsf{f}} =f(\mathcal{\hat{X}}_{\textsf{gen}})$. Our framework is illustrated in Figure \ref{fig:framework}.

\section{NegVerse Data Augmentation  Method}\label{sec:algorithm}

\subsection{\methodName Prompt Format} \label{subsec:prompt_format}

\begin{figure}[t]  
    \centering
    \begin{minipage}{\textwidth} 
    \centering
    \includegraphics[width=1\textwidth]{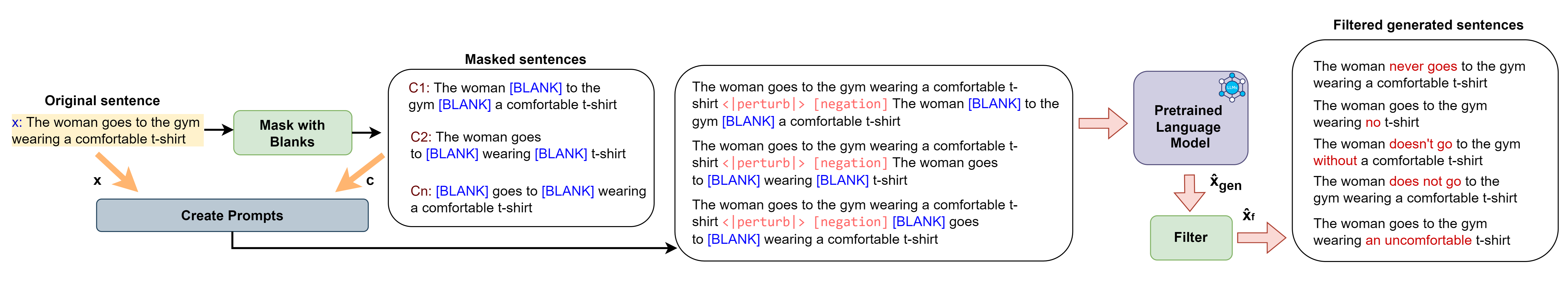}
    \vspace{-0.3in}
    \end{minipage}

    \vspace{0.3cm}  

    \begin{minipage}{0.5\textwidth}  
        \centering
        \includegraphics[width=\textwidth]{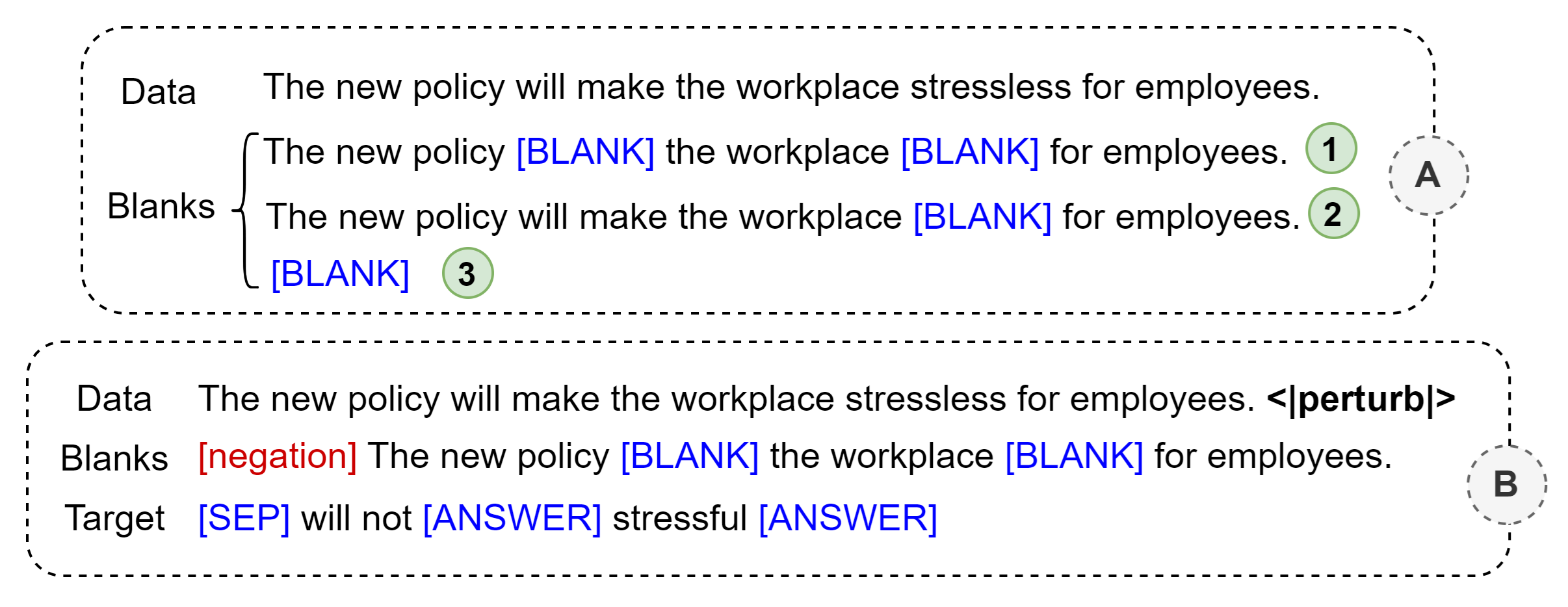}
    \end{minipage}
    \hfill
    \begin{minipage}{0.49\textwidth}  
\centering
\small
\scalebox{0.6}{
\begin{tabular}{cccc}
\toprule
\textbf{{Rule}} & \textbf{Category} & \textbf{POS Tag} & \textbf{Dependency Labels} \\
\midrule
Rule 1 & \textsf{Verbal Negation} & VERB, AUX & Any \\
\midrule
Rule 2 & \textsf{Non-Verbal} & DET & det \\
\midrule
Rule 3 & \textsf{Non-Verbal} & Any & "subj" ,"obj" \\
\midrule
Rule 4 & \textsf{Non-Verbal or Affixal Negation} & ADV & advmod \\
\midrule
Rule 5 & \textsf{Non-Verbal or Affixal Negation} & ADJ & Any \\
\midrule
Rule 6 & \textsf{Non-Verbal} & ADP & prep \\
\bottomrule
\end{tabular} }
    \end{minipage}
    \vspace{-0.15in}
    \caption{\textit{(Top):} Overview of NegVerse steps. The input sentence $x$ is masked with blanks on specific positions based on structural rules. The masked sentences $\bm c$ and the original sentence $x$ are used to create prompts that are fed to a pretrained language model, which then generates $n$ candidate negations, $\hat{\mathcal{X}}_{gen}$. A filtering mechanism selects the most relevant negations from these candidates, producing the final set, $\hat{\mathcal{X}}_f$. \textit{(Bottom Left):} (A) The input data is masked at different spans using the token \textsf{\color{blue}[BLANK]}.  
    Masks cover different parts of the sentence, or the entire sentence (see sentence 3). (B) Training samples concatenate the input text with the masked sentence and the target words needed to fill the blanks. Each span is separated by the token \textsf{\color{blue}[ANSWER]}, and \textsf{\color{blue}[SEP]} separates the context from the answers. During inference, the model accepts the sentence as input, masks the sentence, and predicts the words to fill the blanks, effectively negating the input text. \textit{(Bottom Right)}: Summary of our token selection rules for masking. Tokens are chosen based on Part-of-speech (POS) tags and dependency labels. \methodName masks either the selected token or its entire subtree. }
    
    \vspace{-0.2in}
    \label{fig:framework}
\end{figure}

\paragraph{Prompt Design.} Our model aims to generate negated sentences that meet three key criteria: \textit{closeness}, {\textit{quality}}, and {\textit{diversity}}. 
Closeness ensures the negated sentence minimally differs from the original in structure and meaning.  
Quality emphasizes grammatical correctness, syntactic accuracy, and coherence. Diversity involves generating a variety of negations across different sentence spans, including verbal, non-verbal, and affixal forms, to enrich the dataset and test multiple negation forms. To maintain closeness, we place \textsf{\color{blue}[BLANK]} tokens at likely negation points in the original sentence $x$ and generate various perturbations for each blank. Our prompt format, adapted from Polyjuice, includes a negation control code, a blank sentence, and outputs separated by the \textsf{\color{blue}[ANSWER]} token, allowing the model to generate up to $n$ possible negations per blank. During training, both individual tokens and entire sentences are masked to teach the model sentence structure and negation patterns. We provide details about the metrics for assessment in Section \ref{sec:experiments}.
 
\vspace{-0.09in}
\paragraph{Masking/Blanks Placement Strategy.}  
We propose a masking strategy that enhances negation generation by strategically placing blanks in sentences, addressing limitations in traditional methods like Polyjuice, which often miss key elements such as main verbs, auxiliary verbs, contractions like "\texttt{\color{CadetBlue}wasn't}, and tense variations. Our approach masks key components, including verbs, adjectives, and specific nouns, to support both verbal and non-verbal negations with flexible granularity, allowing for individual token or subtree masking. We developed the token selection rules, summarized in the Bottom Right Table of Figure \ref{fig:framework} based on sentence structure analysis and token functions, covering various aspects of sentence construction like determiners, subjects, objects, adverbs, adjectives, and prepositions, thus enabling the generation of diverse forms of negation. More information about the masking strategy and examples of prompt formats are provided in Appendix~\ref{app:mask_sent}.

\subsection{\methodName Prompt-Tuning Process}\label{sec:finetuning}  

\begin{wraptable}{l}{0.52\textwidth}
\vspace{-0.15in}
\centering
\small 
\scalebox{0.98}{
\begin{tabular}{lc}
\toprule
\textbf{Dataset} & \textbf{Samples \#} \\
\midrule
\textsc{Affixal Negation (SST-2)} & $59$ \\
\textsc{Affixal Negation with Llama 2} & $130$ \\
\textsc{Non-Verbal Negation (NaN-NLI)} & $173$ \\
\midrule
\textbf{Total} & \textbf{$362$} \\
\bottomrule
\end{tabular}}
\vspace{-0.1in}
\caption{Summary of dataset samples used for \methodName across different negation types. }
\vspace{-0.1in}
\label{table:dataset_summary}
\end{wraptable} 

We learn the generator using prompt tuning to update only the virtual token embeddings, while keeping a GPT-2 baseline model frozen. To achieve this, we leverage the non-verbal negations from \textsc{NaN-NLI} \cite{truong-etal-2022-another}, and the affixal negations from \textsc{SST-2} \cite{socher-etal-2013-recursive}, as shown in Table \ref{table:dataset_summary}.  
Given the underrepresentation of affixal negations in the existing datasets, we develop a new dataset of diverse affixal negations from minimal input \cite{app14052074} using one-shot learning and LLaMA2. 
During the prompt tuning process, each sentence is masked either by negated parts or entirely, which helps the model learn to handle different spans and levels of context, thereby enhancing its ability to produce accurate negations. The masked sentences are tokenized, padded, and then split into training and validation sets. 
Additional details on the datasets and hyperparameters are provided in Appendix~\ref{app:train_data} and Appendix~\ref{app:hyperparameter}, respectively.

\subsection{Proposed Filtering Mechanism for Degenerate Sentences}
\begin{wrapfigure}{R}{0.63\textwidth}
\vspace{-0.31in}
{\centering
\begin{minipage}{\linewidth}
\footnotesize

\begin{algorithm}[H]
\caption{\methodName Filtering Mechanism}
\label{alg:select_sentences}
\begin{algorithmic}[1]
\STATE \textbf{Input:} $\{\hat{\mathcal{X}}_{\textsf{gen},i}\}_{i=1}^m$: Generated negated set, $\{x_i\}_{i=1}^m$: affirmative sentences, $\epsilon$: negations sample number, $B=0.5$: Levenshtein distance threshold
\FOR{$i \in [m]$}
\STATE $x_i' \gets \texttt{Trim}(\texttt{Lowercase}(x_i))$
\STATE $\mathcal{X}'_{\textsf{gen},i} \gets \texttt{Trim}(\texttt{Lowercase}(\hat{\mathcal{X}}_{\textsf{gen},i}))$
\STATE $\mathcal{X}_{\tau,i}=\emptyset$
\FOR{$x'_{\textsf{gen}}\in \mathcal{X}'_{\textsf{gen},i}$} 
    \IF{$x'_{\textsf{gen}} \neq$ ""}
        \STATE  $d \gets \texttt{LevenshteinDistance}(x'_{\textsf{gen}},x'_{i})$ 
            \STATE  $\mathcal{X}_{\tau,i} = 
\begin{cases}
    \mathcal{X}_{\tau,i} \cup \{x'_{\textsf{gen}}\}, &  x'_{\textsf{gen}} \notin \mathcal{X}_{\tau,i} \land  d < B  \\
    \mathcal{X}_{\tau,i}, & \text{o.w.}
\end{cases}$ 
    \ENDIF
\ENDFOR

\STATE  $  \hat{\mathcal{X}}_{\textsf{f},i}=\{\hat{x}^{(1)}_{\textsf{f},i},\dots, \hat{x}^{(\epsilon)}_{\textsf{f},i}\} \sim \text{Uni}\big( \texttt{NegBERT}(\mathcal{X}_{\tau,i} ) \big) $   

\ENDFOR    
\STATE \textbf{Output:} Filtered negation sets $\{\hat{\mathcal{X}}_{\textsf{f},i}\}_{i=1}^m$
\end{algorithmic}
\end{algorithm}
 \end{minipage}
\par
}
\end{wrapfigure}
  
Even though our proposed approach is designed to generate fluent and diverse negations, some of the generated outputs may still contain errors or nonsensical phrases. To address this, we propose a filtering process, outlined in Algorithm~\ref{alg:select_sentences}, that normalizes the original and generated sentences by converting it to lowercase and removing trailing punctuation or whitespace (lines 3-5), removes duplicates and uses Levenshtein distance to retain sentences that closely resemble the original (lines 7-10). We use NegBERT to detect the negation cues \cite{khandelwal2020negberttransferlearningapproach}, and we output $\epsilon$ negations, that were uniformly sampled from the set extracted by NegBERT, to increase the diversity in the sets (line 12).

\section{Empirical Results}\label{sec:experiments}

\paragraph{Datasets, Baselines and Metrics.} We evaluate our approach on five datasets: the Stanford Natural Language Inference (SNLI) dataset \cite{bowman-etal-2015-large}, the Semantic Textual Similarity Benchmark (STS) \cite{muennighoff2023mtebmassivetextembedding}, COPA dataset \cite{copa}, and the SemEval Aspect-Based Sentiment Analysis datasets for both restaurant and laptop domains \cite{pontiki-etal-2014-semeval}. We compare \methodName against Polyjuice \cite{wu2021polyjuicegeneratingcounterfactualsexplaining} and evaluate the generated text using (i) Levenshtein Distance (NLD) \cite{nguyen2024llmsgeneratingevaluatingcounterfactuals, ross2021explainingnlpmodelsminimal, treviso2023crestjointframeworkrationalization} that measures the minimal edits required to transform one sentence into another; and (ii) Syntactic Tree Edit Distance (Syntactic), which focuses on surface-level changes \cite{Zhang1989SimpleFA}, to assess closeness. For diversity, we use the Self-BLEU Score \cite{zhu2018texygenbenchmarkingplatformtext}, and for grammaticality and fluency we use a fine-tuned BERT model, following \cite{yoshimura-etal-2020-reference}. The quality of the generated sentences is further evaluated using Perplexity (PPL) \cite{nguyen2024llmsgeneratingevaluatingcounterfactuals,treviso2023crestjointframeworkrationalization}. 
 We provide more details and results, including generation examples and degenerate cases of \methodName, in Appendix \ref{app:errors}.
   
\definecolor{answer}{RGB}{255, 90, 0}  

\definecolor{lightorange}{RGB}{255, 230, 200} 
\definecolor{darkorange}{RGB}{204, 85, 0} 


\begin{table}[h]
    \centering
\scalebox{0.75}{
    \begin{tabular}{ccccccccccc}
        \toprule
        \textbf{Masking }&\textbf{Dataset}& \textbf{Generator} & \multicolumn{2}{c}{\textbf{Closeness}} & \multicolumn{1}{c}{\textbf{Diversity}} & \multicolumn{3}{c}{\textbf{Quality}} \\
        \cmidrule(lr){4-5} \cmidrule(lr){6-6} \cmidrule(lr){7-9}
       \textbf{Type} && & \textbf{NLD} $\downarrow$ & \textbf{Syntactic} $\downarrow$ & \textbf{Self-BLEU} $\downarrow$ & \textbf{Fluency} $\uparrow$ & \textbf{Grammar} $\uparrow$ & \textbf{PPL} $\downarrow$ \\
        \midrule
     \parbox[t]{1mm}{\multirow{10}{*}{\rotatebox[origin=c]{90}{\large Token Level}}}  & SNLI& \texttt{\methodName (ours)}& \textbf{0.200} & \textbf{1.275} & 0.631 & \textbf{0.783} & \textbf{0.814} & \textbf{185.535} \\
        && \texttt{Polyjuice}  & 0.269 & 2.363 & \textbf{0.465} & 0.781 & 0.813 & 249.741 \\
        \cmidrule(lr){2-9} 
        & STS& \texttt{\methodName (ours)} & \textbf{0.216} & \textbf{1.190} & 0.594 & 0.807 & 0.829 & \textbf{295.861} \\
         & & \texttt{Polyjuice}   & 0.306 & 2.360 & \textbf{0.422} & \textbf{0.809} & \textbf{0.831} & 346.224 \\
                \cmidrule(lr){2-9} 
        & COPA& \texttt{\methodName (ours)} & \textbf{0.317} & \textbf{0.824} & 0.415 & \textbf{0.840} & 0.850 & 404.206 \\
        &  & \texttt{Polyjuice}   & 0.434 & 2.451 & \textbf{0.242} & 0.838 & \textbf{0.856} & \textbf{249.493} \\
                \cmidrule(lr){2-9} 
        & Restaurant& \texttt{\methodName (ours)} & \textbf{0.189} & \textbf{1.443} & 0.655 & 0.742 & 0.766 & 141.715 \\
        &   & \texttt{Polyjuice}  & 0.233 & 2.008 & \textbf{0.564} & \textbf{0.743} & \textbf{0.768} & \textbf{134.103} \\
                \cmidrule(lr){2-9} 
        &Laptop & \texttt{\methodName (ours)} & \textbf{0.199} & \textbf{1.490} & 0.629 & \textbf{0.757} & 0.773 & 163.520 \\
        & & \texttt{Polyjuice}   & 0.253 & 2.192 & \textbf{0.530} & 0.756 & 0.773 & \textbf{147.523} \\
       
        \midrule
        \parbox[t]{1mm}{\multirow{10}{*}{\rotatebox[origin=c]{90}{\large Subtree}}} & SNLI& \texttt{\methodName (ours)}  & \textbf{0.200} & \textbf{1.275} & 0.631 & \textbf{0.783} & \textbf{0.814} & \textbf{185.535} \\
       & & \texttt{Polyjuice}   & 0.269 & 2.363 & \textbf{0.465} & 0.781 & 0.813 & 249.741 \\
                \cmidrule(lr){2-9} 
       & STS& \texttt{\methodName (ours)}& \textbf{0.216} & \textbf{1.185} & 0.606 & \textbf{0.809}& \textbf{0.832} & \textbf{296.538} \\
        && \texttt{Polyjuice}   & 0.403 & 3.486 & \textbf{0.328} & 0.823 & 0.845 & 352.290 \\
                \cmidrule(lr){2-9} 
        & COPA& \texttt{\methodName (ours)} & \textbf{0.205} & \textbf{1.300} & 0.640 & \textbf{0.770} & \textbf{0.810} & \textbf{190.654} \\
       & & \texttt{Polyjuice}   & 0.275 & 2.400 & \textbf{0.460} & 0.760 & 0.805 & 250.890 \\
               \cmidrule(lr){2-9} 
       & Restaurant & \texttt{\methodName (ours)} & \textbf{0.206} & \textbf{1.509} & 0.634 & \textbf{0.748} & \textbf{0.772} & \textbf{143.577} \\
        && \texttt{Polyjuice}  & 0.361 & 3.385 & \textbf{0.404} & 0.763 & 0.786 & 259.636 \\
               \cmidrule(lr){2-9} 
       &  Laptop& \texttt{\methodName (ours)} & \textbf{0.216} & \textbf{1.547} & 0.608 & 0.762 & 0.778 & 180.621 \\
        && \texttt{Polyjuice}   & 0.382 & 3.487 & \textbf{0.371} & \textbf{0.780} & \textbf{0.794} & \textbf{152.233} \\
        \bottomrule
    \end{tabular}}
    \caption{Experimental results of \methodName and Polyjuice for token level and subtree masking types using closeness, diversity and quality criteria. The bold numbers indicate the best performance.  
    }
    \label{table:eval_metrics_subtree}
\vspace{-0.25in}
    
\end{table}

\vspace{-0.05in}
\paragraph{Results and Discussion.} We evaluate the performance of our proposed method, \methodName, and the baseline Polyjuice across all datasets using the closeness, diversity, and quality criteria. The results are presented in Table~\ref{table:eval_metrics_subtree}. We observe that \methodName outperforms Polyjuice in closeness and text quality for both token and subtree masking criteria. Our method achieves a lower Levenshtein distance, indicating better lexical similarity to the original sentences, and a lower syntactic score, reflecting better preservation of syntactic structure. In contrast, Polyjuice often introduces unrelated concepts and alters sentence types, affecting coherence, despite offering greater diversity with a lower Self-BLEU score. Moreover, our results show that both models have similar fluency and grammaticality with token masking, but Polyjuice slightly outperforms \methodName in these aspects with subtree masking. This suggests Polyjuice performs better under challenging conditions but it does not necessarily produce more relevant text to the original content. Table \ref{tab:single_exammple} shows an example of negation generation from the two approaches. We expand our discussion and provide more details and results, including generation examples and degenerate cases of \methodName, in Appendix \ref{app:train}.

\begin{table}[h!]
\centering

\scalebox{0.85}{{ 
\begin{tabular}{p{0.3\textwidth} p{0.35\textwidth} p{0.4\textwidth}}
\hline
 &
 \methodName
 & 
 Polyjuice
 \\
\hline
{\textbf{Original}:} They were cooking dinner and serving it to their guests. \vspace{0.15cm} \newline   {\textbf{Masked}:} They {\color{blue}[BLANK]} cooking dinner and serving it to their guests. \vspace{0.1cm}& 
1. \textcolor{gray}{They \highlightres{weren't} cooking dinner and serving it to their guests.} \newline
2. \textcolor{gray}{They \highlightres{were not} cooking dinner and serving it to their guests.} \newline
3. \textcolor{gray}{They \highlightres{didn't care for} cooking dinner and serving it to their guests.} & 
1. \textcolor{gray}{They \highlightres{cook} cooking dinner and serving it to their guests.} \newline
2. \textcolor{gray}{They \highlightres{cook in the kitchen and not the} \highlightres{dining room because the dining room is} \highlightres{farthest from} cooking dinner and serving it to their guests.} \\ \hline
\end{tabular}} }
\caption{A negation generation example for \methodName and Polyjuice. {\color{blue}[BLANK]} marks the masked parts of the original sentence, and the highlighted text shows the generated fill-ins. \methodName produces outputs that closely mirror the original sentence, while Polyjuice offers more variety in outputs, which contributes to diversity, but can compromise the relevance and fidelity of the text.}
\vspace{-0.2in}\label{tab:single_exammple}
\end{table}

\section{Conclusions}\label{sec:conclusions}

In this work, we focus on improving the robustness of LLMs robustness on negated statements by proposing \methodName, a method capable of generating various types of negations, including verbal, non-verbal, and affixal. We provide new masking rules and propose a filtering mechanism to identify negation cues and remove degenerate examples, producing diverse and in parallel meaningful negated sentences.
We experiment with five real-world datasets and \methodName outperforms existing methods and generates negations with higher lexical similarity to the original sentences, better syntactic preservation, and greater negation diversity. Our empirical results also highlight that the proposed approach can generate negated sentences without specific guidance on blank placement. \vspace{-0.05in}

\paragraph{Limitations and Future Work.} While \methodName excels in preserving syntactic structure and offers a greater variety of negation forms, it still produces some degenerate outputs, particularly when blanks are placed at the end of sentences, leading to grammatically correct but contextually meaningless results. Furthermore, although \methodName generates a range of affixal negations, certain expected forms are missing. Finally, automated and accurate annotation is essential for the generated negations, as negations can either preserve or invert labels depending on the task.

\newpage

\bibliographystyle{plain}
\bibliography{main.bib}


\appendix
\newpage
\section*{Supplemental material}


\section{Types of Negations.}
\begin{table}[ht!]
\centering
\small 
{ \begin{tabular} 
 {@{}p{0.4\textwidth} p{0.55\textwidth}@{}}
\toprule
\textbf{Negation Type} & \textbf{Examples} \\
\midrule
\parbox[t]{0.38\textwidth}{{\textbf{Verbal Negation}}\\ {(Syntactic type)}\\ \texttt{\color{CadetBlue}not, n’t, didn’t, cannot, won’t, etc.}} & They are still \highlight{not} integrated into the German community. \par
\vspace{1.mm}
We \highlight{didn't} go to the beach because it started raining. \par
\vspace{1.mm}
She \highlight{won’t} be attending the meeting. \\
\midrule
\parbox[t]{0.38\textwidth}{{\textbf{Non-Verbal\\ Negation}}\\ {(Syntactic type)}\\ \texttt{\color{CadetBlue}no, nothing, nowhere, nobody, none, without, etc.}} & I have \highlight{no} doubt that we will reach our goal. \par
\vspace{1.mm}
He found \highlight{nothing} in the drawer. \par
\vspace{1.mm}
The lost keys were found \highlight{nowhere} in the house. \par
\vspace{1.mm}
He completed the task \highlight{without} any help. \\
\midrule
\parbox[t]{0.38\textwidth}{{\textbf{Affixal Negation}}\\ {(Morphological type)}\\ \texttt{\color{CadetBlue}un-, in-, dis-, -less, non-, etc.}} & Her reaction was \highlight{unexpected} given the circumstances. \par
\vspace{1.mm}
She felt \highlight{hopeless} after repeated failures. \par
\vspace{1.mm}
The product was \highlight{non-existent} on the shelves. \\
\bottomrule
\end{tabular}}

\caption{Overview of verbal, non-verbal, and affixal negation forms, with corresponding examples demonstrating their application in sentences.}
\label{table:negation}
\end{table}
\vspace{-0.1in}
There are two main types of negations: morphological and syntactic negations, as outlined in Table \ref{table:negation}. Morphological negations create negative expressions by adding affixes to words, either as prefixes or suffixes. A prefixal negation adds prefixes to the beginning of words and includes common prefixes like \texttt{\color{CadetBlue}un-} (e.g., \textit{unhappy}), \texttt{\color{CadetBlue}in-/im-/il-/ir-} (e.g., \textit{inaccurate}, \textit{impossible}, \textit{illegal}, \textit{irrelevant}), \texttt{\color{CadetBlue}dis-} (e.g., \textit{disagree}), and \texttt{\color{CadetBlue}non-} (e.g., \textit{nonexistent}). A suffixal negation adds suffixes to the end of words and includes the common suffix \texttt{\color{CadetBlue}-less} (e.g., \textit{hopeless}, \textit{meaningless}). These affixes alter the meaning of the base words to convey negation, absence, or opposition \cite{van-son-etal-2016-building}. Syntactic negations utilize grammatical structures and specific words to negate a sentence. This typically includes negative particles like \texttt{\color{CadetBlue}not} and \texttt{\color{CadetBlue}no} (e.g., \textit{She is not coming}; \textit{There is no water}), negative pronouns like \texttt{\color{CadetBlue}nobody} and \texttt{\color{CadetBlue}nothing} (e.g., \textit{Nobody knows}; \textit{Nothing happened}), negative adverbs like \texttt{\color{CadetBlue}never} and \texttt{\color{CadetBlue}nowhere} (e.g., \textit{She never comes}; \textit{They went nowhere}), negative determiners like \texttt{\color{CadetBlue}no} and \texttt{\color{CadetBlue}neither} (e.g., \textit{No students passed}; \textit{Neither option is good}), and negative conjunctions like \texttt{\color{CadetBlue}nor} and \texttt{\color{CadetBlue}neither...nor} (e.g., \textit{She didn't call, nor did she email}; \textit{Neither he nor his friends came}).
\section{Prompt Design}\label{app:mask_sent}

In this section, we provide further details on our six rules of the masking strategy, which were outlined earlier in Section \ref{subsec:prompt_format}.

\begin{figure}[h]
    \centering
    \includegraphics[width=0.8\textwidth]{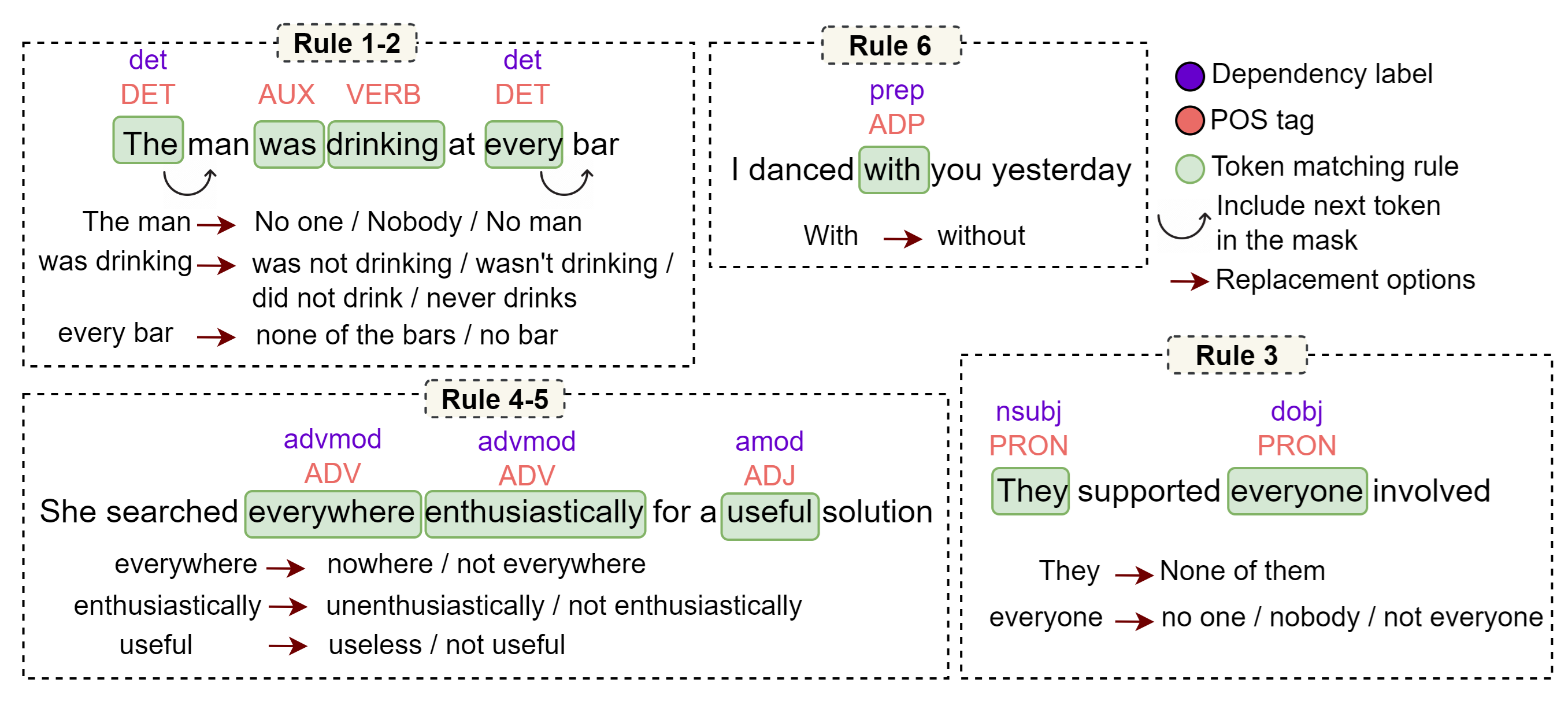}\vspace{-0.2in}
    \caption{Illustrative example of sentences that follow our proposed blank placement rules. Although some sentences comply with multiple rules, only the words matching the specific rule are highlighted in green for each case. Below each sentence, possible negations that can be introduced by filling in the blanks are provided. This example demonstrates how this placement strategy can produce diverse forms of negation. The arrow sign ($\rightarrow$) indicates that when the word is a determiner (DET), it masks the accompanying noun or adjective, allowing the model to generate richer negations.}
    \label{fig:rule_example}
    \end{figure}

\paragraph{Rule 1:} The first rule targets verbal negations by selecting verbs (\texttt{VERB}) and auxiliaries (\texttt{AUX}) for masking, as these are key components in forming negations. For instance, in the sentence "\textit{She was eating an apple}", masking "\texttt{\color{CadetBlue}was}" and "\texttt{\color{CadetBlue}eating}" allows the model to generate the negation "\textit{She was not eating an apple.}" This approach effectively negates the core actions or states in the sentence, as illustrated in Figure~\ref{fig:rule_example}.
\paragraph{Rule 2:} The second rule targets non-verbal negation by focusing on determiners (\texttt{DET}) with the dependency label \texttt{det}. Determiners like "\texttt{\color{CadetBlue}the}", "\texttt{\color{CadetBlue}a}", and "\texttt{\color{CadetBlue}an}" are crucial for defining noun phrases. The model selects a determiner and the following token for transformation, such as changing "\texttt{\color{CadetBlue}the man}" to "\texttt{\color{CadetBlue}no one}", as shown in Figure~\ref{fig:rule_example}. Unlike Rule 3, which may negate entire phrases, Rule 2 specifically alters the determiner. Other examples include:

{\highlightap{Peter wanted \redout{some} part of it. $\rightarrow$ Peter wanted \highlight{none} of it.}}

\textbf{Rule 3:} This rule focuses on negating objects ("obj") and subjects ("subj"), which are essential for defining who is performing an action and what is being acted upon. Negating the subject (`"subj"`) changes who or what is performing the action. For instance:

{\highlightap{\redout{They} will attend the meeting $\rightarrow$ \highlight{No one} will attend the meeting}} \\
{\highlightap{\indent \redout{They} will attend the meeting $\rightarrow$ \highlight{None of them} will attend the meeting}}

Negating the object (`"obj"`) changes what is being acted upon, affecting the outcome of the action. For example:

{\highlightap{She found \redout{the key} $\rightarrow$ She found \highlight{nothing}}} \\
{\highlightap{\indent She went to \redout{the gym} → She went to  \highlight{no gym}}}

\paragraph{Rule 4:} This rule targets adverbs (\texttt{ADV}) with the dependency label \texttt{advmod} for non-verbal or affixal negation. By masking adverbs, the rule generates various negations, such as changing "\texttt{\color{CadetBlue}everywhere}" to "\texttt{\color{CadetBlue}nowhere}" or "\texttt{\color{CadetBlue}not everywhere}", and "\texttt{\color{CadetBlue}enthusiastically}" to "\texttt{\color{CadetBlue}unenthusiastically}" or "\texttt{\color{CadetBlue}not enthusiastically}". This approach modifies the action’s scope or intensity and incorporates morphological changes.

\paragraph{Rule 5:} This rule enables non-verbal or affixal negation by targeting adjectives (\texttt{ADJ}), allowing for direct negation or morphological changes. For example, "\textit{The solution is useful}" can be transformed to "\textit{The solution is useless}" (affixal) or "\textit{The solution is not useful}" (non-verbal).

\paragraph{Rule 6:}This rule handles non-verbal negation by targeting prepositions (\texttt{ADP}) that provide context such as location or time. It is used less frequently and only when the mask subtree is active, due to its limited variations. For example,

{\highlightap{She will meet us \redout{at the restaurant} $\rightarrow$ She will meet us \highlight{nowhere}}}

\begin{figure}[h]
    \centering
    \includegraphics[width=0.9\textwidth]{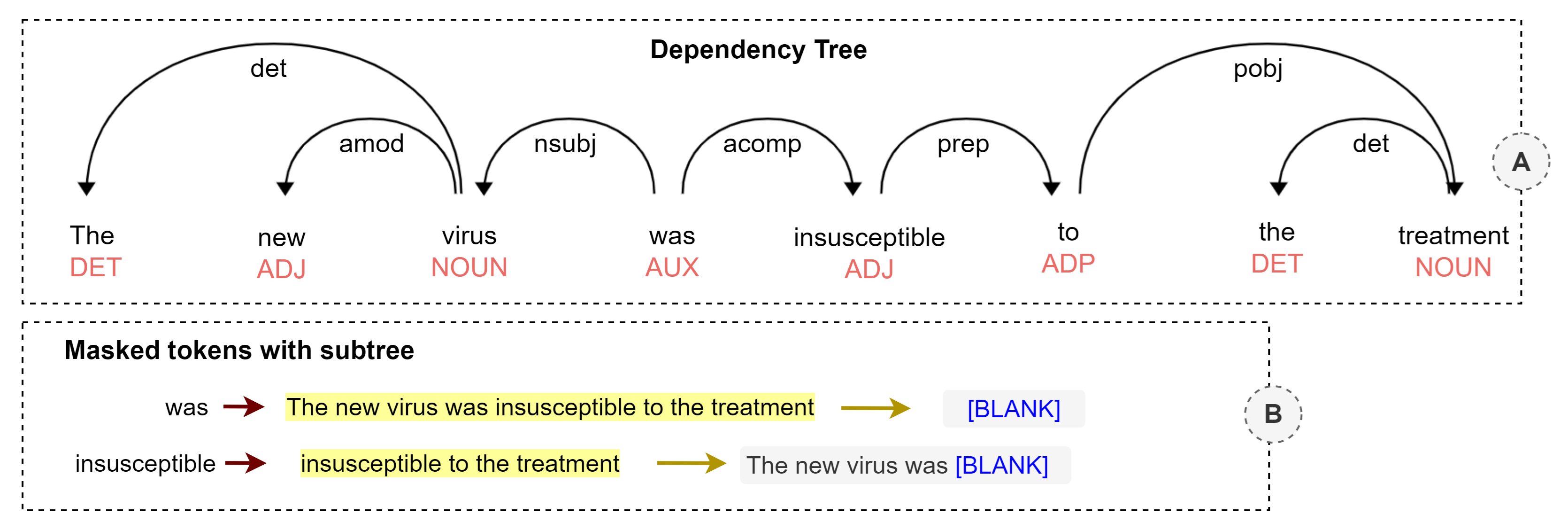}
    \vspace{-0.1in}
    \caption{Dependency parse tree representing the grammatical structure of an example sentence. (A) The syntactic structure of the sentence, with arcs representing grammatical dependencies between words. Dependency labels (dep tags) are displayed on the arcs, and part-of-speech tags (POS) are shown under each word, illustrating the sentence's syntactic structure. (B) Tokens within the subtree rooted at the selected token are highlighted in yellow. The highlighted tokens are then masked with {\color{blue}[BLANK]} instead of just the individual token. If a verb is selected, all words dependent on it within the sentence are included in the subtree, resulting in the entire sentence being masked.}

    \label{fig:subtree}
    \end{figure}

\begin{figure}[h]
    \centering
    \includegraphics[width=0.6\textwidth]{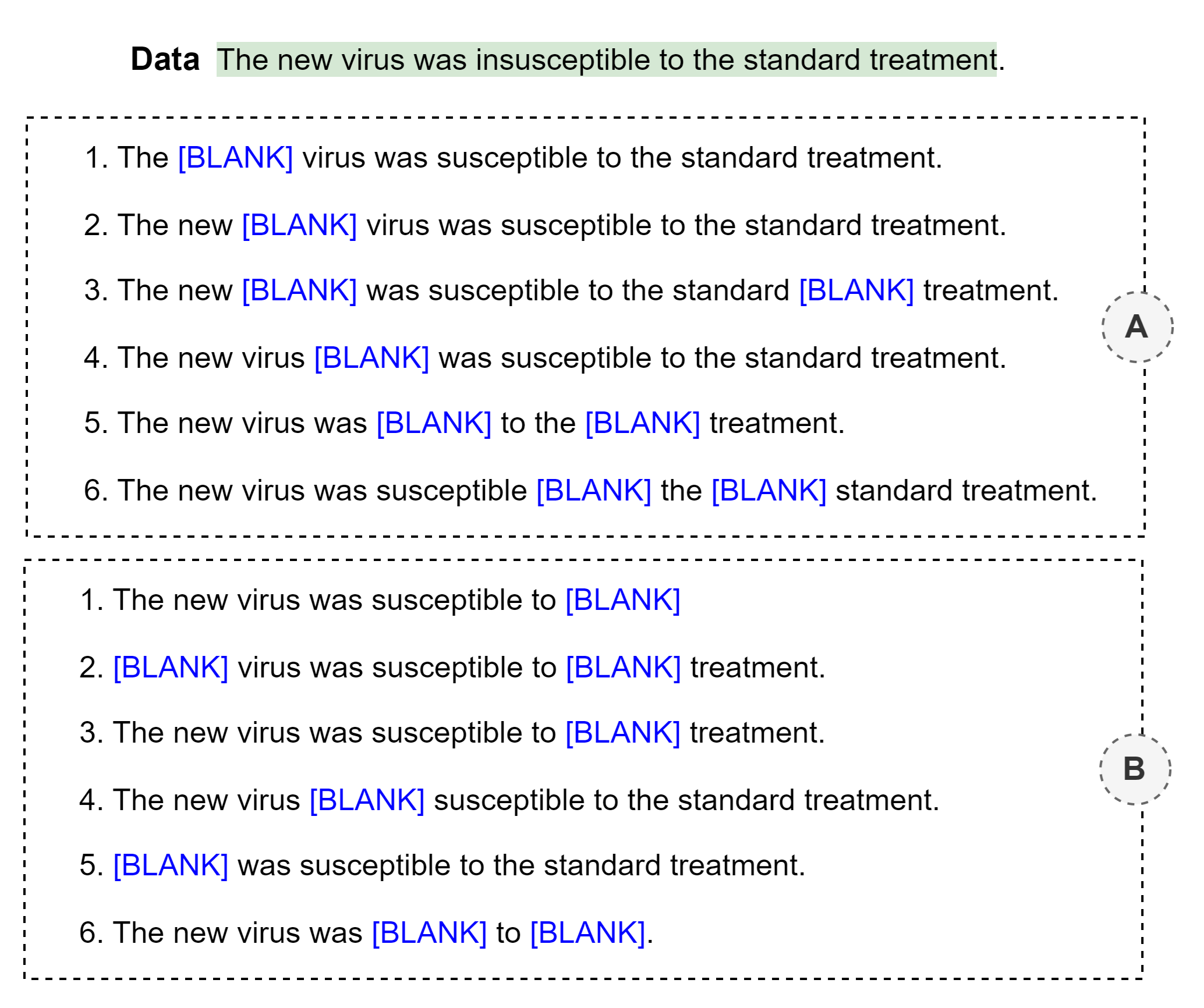}
    \vspace{-0.1in}
    \caption{ An illustrative example of sentence masking. The masking function considers a maximum of two tokens per sentence, and six different masked sentences. Part (A) represents the masked sentence with Polyjuice automatic masking, where the main verb is masked in none of the options, nor were the adjectives with possible affixal negated forms like "insusceptible". Additionally, in Option $3$, a {\color{blue}[BLANK]} was inserted rather than replacing a token. Part (B) shows how the proposed approach masks a sentence.  In particular, Option $6$ masks the adjective, Option $4$ masks the main verb, and the other options mask in places to produce non-verbal negations.}
    \label{fig:mask_example}
    \end{figure}

We provide an example illustrating the impact of the six proposed rules in Figure \ref{fig:mask_example}. We also show the broader context included in the negation from a subtree's selected token and syntactic dependents \cite{Shen2014DependencyPR} in Figure~\ref{fig:subtree}.


\section{Additional Experimental Setup Details and Results}
\label{app:train}
\subsection{Training Data Details}
\label{app:train_data}

\begin{table}[h]
\centering
\small 
{ 
\begin{tabular}{@{}p{4cm} p{4cm} p{5cm}@{}}
\toprule
\textbf{Construction Type} & \textbf{Definition} & \textbf{Example} \\
\midrule
Not + quantifiers &  Not combined with a quantifier, (e.g., Not all, not every, not many, not much). &  \highlight{Not one} person supported  the proposal. \\
\midrule
Not + focus particles &  "Not even" denotes clausal negation; "not only" indicates sub-clausal negation with a positive tone. &  \highlight{Not even} Ed approved of the plan. \\
\midrule
Not + degree expressions &  Marks sub-clausal negation by reducing the intensity of adjectives or adverbs (e.g., "not very confident"). & It somehow sounded \highlight{not quite} right. \\
\midrule
Not + affixal &  Affixal negations of adjectives and adverbs. & It was a \highlight{not undistinguished} private university with a large endowment. \\
\midrule
Not in coordination &  "Not" in a coordinative construction negates only one part of the conjunction, indicating sub-clausal negation. & They are now leaving \highlight{not
on Friday but on Saturday} \\
\midrule
Not in verbless subordinate clauses & "Not" can negate only the verbless subordinate clause & We need someone \highlight{not afraid} of taking risks. \\
\midrule
Not in implicit propositions with that & Denies something anticipated or implied in the context & There are spare blankets in here, \highlight{not that} you'll have any need of them.\\
\midrule
Absolute negators & Indicates complete non-existence within a prepositional phrase (e.g., no, never) & They were
friends in \highlight{no time}.\\
\midrule
Approximate negators & Suggests near-zero frequency with a positive implication (e.g., rarely, seldom) & She \highlight{rarely} goes out these days.\\
\bottomrule
\end{tabular}}
\caption{Definitions and examples of different negation types within the \textsc{NaN-NLI} dataset. The highlighted text in each example illustrates the specific negation construction being discussed \cite{truong-etal-2022-another}.}
\label{table:dataset_explanation}
\end{table}

\begin{table}[h]
\centering
\small{ 
\begin{tabular}{p{5cm} p{5cm} c}
\toprule
\textbf{Original Sentence} & \textbf{Negated Sentence} & \textbf{} \\
\midrule
The water in the lake was \redout{pure}, making it \redout{safe} for drinking. & The water in the lake was \highlight{impure}, making it \highlight{unsafe} for drinking. & \textcolor{green}{\scalebox{1.5}{$\checkmark$}} \\
\midrule
The employee's work was \redout{worthy} of the bonus due to the \redout{exceptional effort}. & The employee's work was \highlight{unworthy} of the bonus due to the \highlight{lack of effort}. & \textcolor{green}{\scalebox{1.5}{$\checkmark$}} \\
\midrule
The new employee's \redout{enthusiasm and willingness to learn} made it \redout{easy} for him to receive the necessary support from his colleagues. & The new employee's \highlight{lack of experience} made it \highlight{difficult} for him to receive the necessary support from his colleagues. & \textcolor{red}{\scalebox{1.5}{$\text{x}$}} \\
\midrule
The new employee \redout{quickly connected} with his colleagues \redout{and became an integral part of the team}. & The new employee \highlight{struggled to connect} with his colleagues \highlight{due to his shyness}. & \textcolor{red}{\scalebox{1.5}{$\text{x}$}} \\
\bottomrule
\end{tabular}}
\caption{Comparison of negated and original sentences generated with Llama 2 to illustrate affixal negations examples for training. The generated data were manually analyzed for validity, where sentences that did not correctly convey affixal negations were eliminated from the dataset. Sentences with substantial word substitutions were also excluded, as the goal is to have samples with minimal changes. Parts of the original sentences that were eliminated are crossed out, while the validity of the changes is indicated by \textcolor{green}{\scalebox{1.5}{$\checkmark$}} for correct pairs and \textcolor{red}{\scalebox{1.5}{$\text{x}$}} for incorrect ones.}
\label{table:sentence_comparison}
\end{table}

In section \ref{sec:finetuning}, we provided information about the tuning process of NegVerse by combining the non-verbal negations from \textsc{NaN-NLI} \cite{truong-etal-2022-another}, and the affixal negations from \textsc{SST-2} \cite{socher-etal-2013-recursive} and the new dataset we generated using LLaMA2. In what follows, we provide more information about these datasets.

\paragraph{\textsc{NaN-NLI}:} This dataset is used to evaluate models' capabilities in understanding and processing sub-clausal negation instances in natural language applications. Sub-clausal negation occurs within a clause, rather than negating the entire clause itself. The dataset annotates various aspects of negation, including verbal vs. non-verbal, analytic vs. synthetic, and clausal vs. sub-clausal negation types. Additionally, it captures the constructions used in negation instances, as well as the operations applied to construct hypotheses \cite{truong-etal-2022-another}. The dataset provides a list of construction types used in negation instances, where most cases involve non-verbal negations, as shown in Table~\ref{table:dataset_explanation}.

\paragraph{\textsc{SST-2}:} This dataset is a collection of movie reviews classified as negative or positive. It includes two types of negations: syntactic (SYN) and morphological (AFFIX) \cite{hossain2022analysis}. For training the model with this data, only the AFFIX annotations were filtered, where the negation cue could be translated to a positive sentiment. For example, "\texttt{\color{CadetBlue}unpleasant}" can be translated to "\texttt{\color{CadetBlue}pleasant}". The sentences were converted to positive by applying manual rules, considering cases where the negation cue starts with "\texttt{\color{CadetBlue}un}" or ends with "\texttt{\color{CadetBlue}less}", using a dictionary of affixal negations from \cite{van-son-etal-2016-building}. For instance:

{\highlightap{\textbf{\color{black}Original:} The film is quiet, threatening, and \redout{unforgettable}}}.\\
{\highlightap{\textbf{\color{black}Negated:} The film is quiet, threatening, and \highlight{forgettable}}}.

\begin{figure}[h]
\centering
\includegraphics[width=0.55\textwidth]{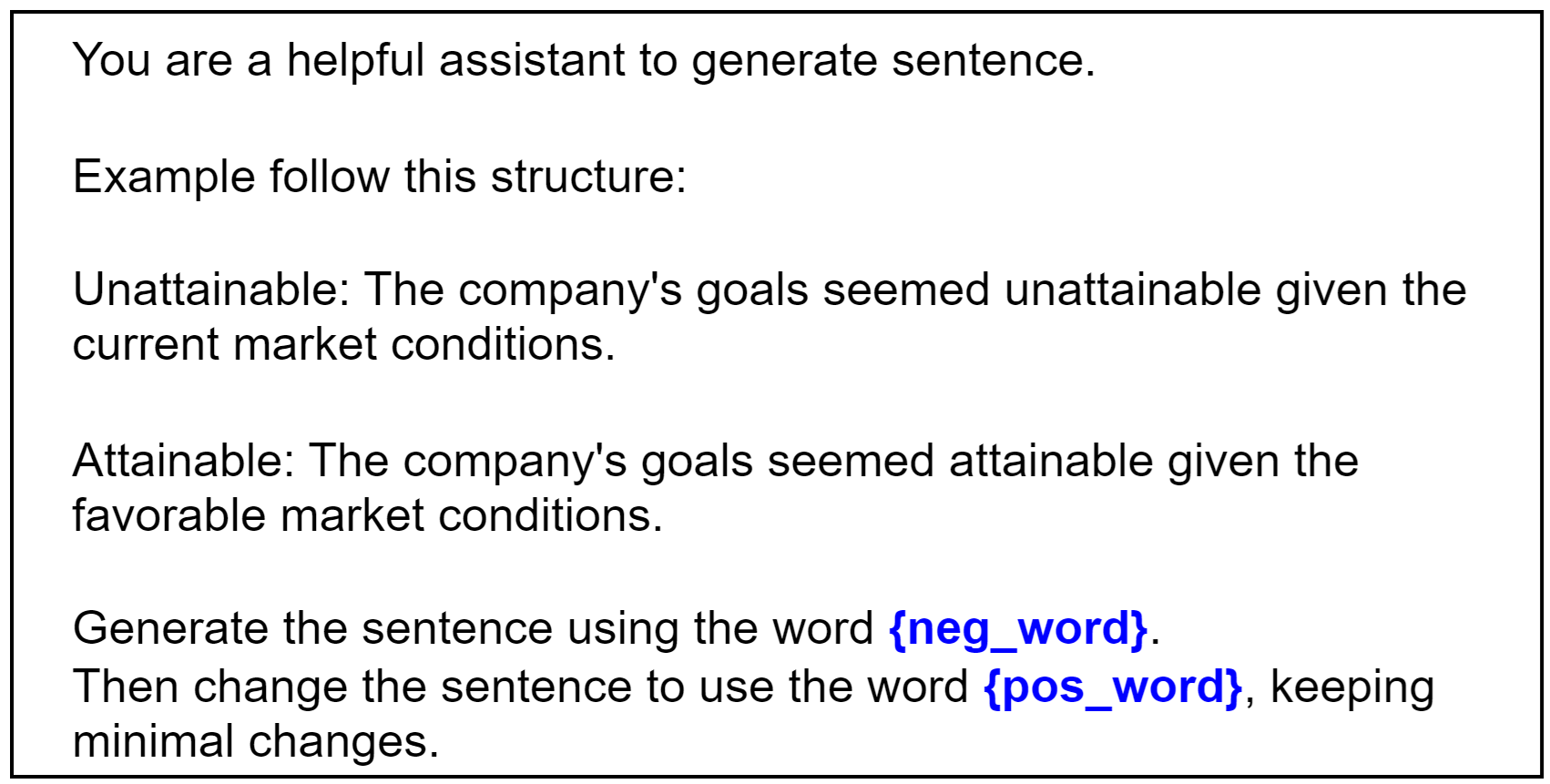}
\caption{Prompt template used to generate data with affixal negations by leveraging one-shot learning with an instruction-following LLM assistant using Llama-2-7b-Chat. The text in blue indicates where the new pair of words is inserted for inference.}
\label{fig:template}
\end{figure}

\paragraph{New dataset: Affixal Negations Generated from LLaMA2.} 
Affixal negations, using prefixes or suffixes, were underrepresented in existing datasets. To address this, we created a new dataset with additional sentence pairs focused on affixal negations using the Llama-2-7b-Chat model and one-shot learning, enabling efficient generation of diverse examples from minimal input \cite{app14052074}. 

We used prompt engineering to guide the model in generating and modifying sentences. The prompts provided structured examples of affixal negations and their transformation into positive forms with minimal changes, as shown in Figure~\ref{fig:template}. For instance, the model replaced "{\highlightap{unattainable}}" with "{\highlightap{attainable}}" in a sentence. Table~\ref{table:sentence_comparison} shows that the model occasionally failed to make minimal changes or correctly apply affixal negations, resulting in the exclusion of such cases from the training dataset. 

The Llama-2-7b-Chat model \cite{touvron2023llama2openfoundation} was selected for generating affixal negation sentences from limited training data due to its ability to produce coherent, contextually accurate text with minimal errors and its strong performance in one-shot learning \cite{chatterjee2024languagemodelsexploitcrosstask, toukmaji2024fewshotcrosslingualtransferprompting}. 
Additionally, its use is cost-free, in contrast to models such as GPT-3 \cite{brown2020languagemodelsfewshotlearners} and GPT-4 \cite{openai2024gpt4technicalreport}.

\subsection{ Hyperparameters }
\label{app:hyperparameter}

We train the model for 31 epochs using the AdamW optimizer, which integrates weight decay directly into the optimization process. The learning rate is set to \(2.5 \times 10^{-2}\) to strike an optimal balance between training efficiency and convergence speed. A weight decay of \(1 \times 10^{-3}\) is employed to address overfitting by penalizing large weights, while a batch size of 16 ensures stable gradient updates within memory constraints. Additionally, 24 virtual tokens are used to prompt-tune the model, allowing for focused adaptation on the specific task. Various hyperparameters were tested and monitored through the learning curve analysis, with these settings yielding the best results in terms of stability and performance. Out of the total 124,458,240 parameters in the pre-trained language model, only 18,432 are trainable, all coming from the virtual token embeddings. This represents just 0.0148\% of the model’s parameters, demonstrating the efficiency of the soft-tuning approach.

\subsection{Software and Hardware }
\label{app:sw_hw}
The proposed algorithms and experiments are implemented in Python, utilizing the PyTorch library. The experiments were conducted using a single NVIDIA Tesla A100 GPU.
The official implementation of NegBERT can be found at \url{https://github.com/adityak6798/Transformers-For-Negation-and-Speculation}.
\subsection{Additional Results}
\label{app:errors}

\begin{table}[h!]
\centering

\scalebox{0.92}{{ 
\begin{tabular}{p{0.25\textwidth} p{0.38\textwidth} p{0.4\textwidth}}
\hline
 & \vspace{0.01in}\textbf{\texttt{\large \methodName}}\vspace{0.05in} & \vspace{0.01in}\textbf{\texttt{\large Polyjuice}} \vspace{0.05in}\\
\hline
{\textbf{Original}:} They were cooking dinner and serving it to their guests. \vspace{0.15cm} \newline   {\textbf{Masked}:} They {\color{blue}[BLANK]} cooking dinner and serving it to their guests. \vspace{0.1cm}& 
1. \textcolor{gray}{They \highlightres{weren't} cooking dinner and serving it to their guests.} \newline
2. \textcolor{gray}{They \highlightres{were not} cooking dinner and serving it to their guests.} \newline
3. \textcolor{gray}{They \highlightres{didn't care for} cooking dinner and serving it to their guests.} & 
1. \textcolor{gray}{They \highlightres{cook} cooking dinner and serving it to their guests.} \newline
2. \textcolor{gray}{They \highlightres{cook in the kitchen and not the} \highlightres{dining room because the dining room is} \highlightres{farthest from} cooking dinner and serving it to their guests.} \\
\hline
{\textbf{Original}:}  Everybody loves the coffee in London.  \vspace{0.15cm} \newline {\textbf{Masked}:} {\color{blue}[BLANK]} \vspace{0.1cm}  & 
1. \highlightres{Nobody loves the coffee in London.} \newline
2. \highlightres{Nobody hates the coffee in London.} & 
1. \highlightres{What is the last name of the person} \highlightres{that Vickers breaks up with?} \newline
2. \highlightres{What is the full name of the person} \highlightres{who has a brother named "Doc"?} \\
\hline
{\textbf{Original}:} The gourmet dinner was delicious and expensive. \vspace{0.15cm} \newline {\textbf{Masked}:} The gourmet dinner was {\color{blue}[BLANK]} and {\color{blue}[BLANK]}. \vspace{0.1cm} & 
1. \textcolor{gray}{The gourmet dinner was \highlightres{unappealing} and \highlightres{not expensive}.} \newline
2. \textcolor{gray}{The gourmet dinner was \highlightres{unappealing} and \highlightres{expensive}.} \newline
3. \textcolor{gray}{The gourmet dinner was \highlightres{unappealing} and \highlightres{not expensive}.} & 
\textcolor{gray}{The gourmet dinner was \highlightres{served in a} \highlightres{t-glass rather than a glass, because the} \highlightres{t-glass was better} and \highlightres{tastier}.} \\
\hline
{\textbf{Original}:} He stayed at the hotel. \vspace{0.15cm} \newline {\textbf{Masked}:} He stayed {\color{blue}[BLANK]}.\vspace{0.1cm} & 
1. \textcolor{gray}{He stayed \highlightres{not at the hotel.}} \newline
2. \textcolor{gray}{He stayed \highlightres{not at the hotel.}} \newline
3. \textcolor{gray}{He stayed \highlightres{away from the hotel.}} & 
1. \textcolor{gray}{He stayed \highlightres{in Germany for three years} \highlightres{before moving back with his family to} \highlightres{Japan.}}  \\
\hline
\end{tabular}} }
\caption{Examples of negation outputs for \methodName and Polyjuice showing differences in closeness, quality and diversity. The {\color{blue}[BLANK]} marks the masked parts of the original sentence, with the highlighted text showing the generated fill-ins. \methodName typically produces outputs that closely mirror the original sentence, maintaining coherence. In contrast, Polyjuice offers more varied outputs, which, while contributing to diversity, sometimes compromise relevance and fidelity to the source text.}
\label{tab:negation_comparison_clos}
\end{table}

\paragraph{Negation Comparisons} In Table~\ref{tab:negation_comparison_clos}, we provide illustrative examples of how \methodName and Polyjuice handle verbs and sentence masking differently. When a verb is masked, \methodName generates various forms of negations, including both contracted and uncontracted versions. In contrast, Polyjuice often introduces unrelated concepts such as "\texttt{\color{CadetBlue}dining room}", "\texttt{\color{CadetBlue}Germany}" and "\texttt{\color{CadetBlue}t-glass}", which are not present in the original sentence and disrupt its overall coherence. Additionally, when entire sentences are masked, \methodName typically produces outputs that closely resemble the original, while Polyjuice frequently creates entirely different sentences, sometimes altering the sentence type altogether, such as changing an affirmative statement into a question, compromising this way the coherence and relevance of the output.

\begin{table}[h]
\centering
\scalebox{0.83}{{ 
\renewcommand{\arraystretch}{1.2} 
\begin{tabular}{p{6cm} p{5cm}ccc}
\hline
& \textbf{\texttt{\large \methodName}} & \textbf{\texttt{\large Gramm.}} & \textbf{\texttt{\large Flu.}} & \textbf{\texttt{\large PPL}} \\ \hline

\multirow{2}{6cm}{{\textbf{Original}:} Her sweater is comfortable and pretty. \vspace{0.2cm} \newline
{\textbf{Masked}:} Her sweater is {\textcolor{blue}{[BLANK]}} and pretty.} & 

1. \textcolor{gray}{Her sweater is \highlightres{uncomfortable} and pretty} & 0.894 & 0.834 & 856.522 \\ 
& 2. \textcolor{gray}{Her sweater is \highlightres {not comfortable}  and pretty} & 0.834 & 0.758 & 406.590 \\ \hline
 
\multirow{2}{6cm}{{\textbf{Original}:} He offers a rational explanation for his decision. \vspace{0.2cm} \newline
{\textbf{Masked}:} He {\textcolor{blue}{[BLANK]}} a rational explanation for his decision.} & 

1.  \textcolor{gray}{He \highlightres{doesn't offer} a rational explanation for his decision.} & 0.961 & 0.964 & 26.859 \\ 
& 2.  \textcolor{gray}{He \highlightres{lacks} a rational explanation for his decision.} & 0.974 & 0.969 & 81.216 \\ \hline

\multirow{2}{6cm}{{\textbf{Original}: A group of kids plays in the spray of water from a fountain.)}\vspace{0.2cm} \newline
{\textbf{Masked}:} \textcolor{blue}{[BLANK]} plays in the spray of water from a fountain.} & 
1. \textcolor{gray}{\highlightres{Not a group of kids} plays in the spray of water from a fountain.} & 0.770 & 0.697 & 84.193 \\ 
 & 2. \textcolor{gray}{\highlightres{None of the kids} plays in the spray of water from a fountain.} & 0.800 & 0.749 & 80.611 \\ \hline
 
\multirow{2}{6cm}{{\textbf{Original}: She spent the day wearing an unique sweater}\vspace{0.2cm} \newline
{\textbf{Masked}:} She spent the day wearing {\textcolor{blue}{[BLANK]}} sweater} & 
1. \textcolor{gray}{She spent the day wearing \highlightres{no} sweater} & 0.939 & 0.877 & 596.420 \\ 
 & 2. \textcolor{gray}{She spent the day wearing \highlightres{nothing} sweater} \textcolor{red}{\scalebox{2}{$\text{x}$}} & 0.956 & 0.956 & 979.119 \\ \hline
 
\multirow{2}{6cm}{{\textbf{Original}: The cat slept peacefully in the sun.} \vspace{0.2cm} \newline
{\textbf{Masked}:} {\textcolor{blue}{[BLANK]}} slept peacefully in {\textcolor{blue}{[BLANK]}}.} & 

1. \textcolor{gray}{\highlightres{The cat did not} slept peacefully in \highlightres{the sun}.} \textcolor{red}{\scalebox{2}{$\text{x}$}} & 0.790 & 0.712 & 381.641 \\ 
 & 2. \textcolor{gray}{\highlightres{The cat didn't} slept peacefully in \highlightres{the sun}.} \textcolor{red}{\scalebox{2}{$\text{x}$}}  & 0.861 & 0.782 & 177.690 \\ \hline
 
\end{tabular}} }
\caption{Quality evaluation of various generated sentences. This table compares sentences generated by \methodName with their original counterparts, showing metrics for grammaticality (Gramm.), fluency (Flu.), and perplexity (PPL). The examples highlight how different word choices and constructions affect these metrics, and reveal insights into the model's performance and limitations in maintaining naturalness and coherence. \textcolor{red}{X} indicates sentences that are grammatically incorrect.}
\label{tab:evaluation}
\end{table}

\begin{table}[h]
\centering
\scalebox{0.8}{{ 
\renewcommand{\arraystretch}{1.2} 
\begin{tabular}{p{6cm}p{5cm}ccc}
\hline

& \textbf{\texttt{\large \methodName}} & \textbf{\texttt{\large Gramm.}} & \textbf{\texttt{\large Flu.}} & \textbf{\texttt{\large PPL}} \\ \hline

\multirow{2}{6cm}{{\textbf{Original}: He offers a rational explanation for his decision.}\vspace{0.2cm} \newline
{\textbf{Masked}:} He offers a rational explanation for {\textcolor{blue}{[BLANK]}}.} &
 
1. \textcolor{gray}{He offers a rational explanation for \highlightres{\texttt{|> [|> [|> [|> [|> [|> [|> }}} \highlightres{\texttt{[|> [|> [|> [|> [|>}}  & 0.769 & 0.760 & 3.511 \\ 
 & 2. \textcolor{gray}{He offers a rational explanation for \highlightres{his decision.}} & 0.976 & 0.980 & 44.482 \\ \hline

\multirow{2}{6cm}{{\textbf{Original}: a young woman fishing off a dock at sunset.}\vspace{0.2cm} \newline
{\textbf{Masked}:} A young woman fishing off a dock at {\textcolor{blue}{[BLANK]}}.} &
1. \textcolor{gray}{A young woman fishing off a dock at \highlightres{a young woman fishing off a dock} \highlightres{at dusk......... not a young woman}} \highlightres{........ a young woman..} & 0.822 & 0.812 & 16.623 \\ 
 & 2. \textcolor{gray}{A young woman fishing off a dock at \highlightres{no sunset.}} & 0.737 & 0.688 & 550.823 \\ \hline

\multirow{1}{6cm}{{\textbf{Original}: A bald headed man in business casual attire is amused by something happening off-screen.}\vspace{0.2cm} \newline
{\textbf{Masked}:} A bald headed man in business casual attire {\textcolor{blue}{[BLANK]}} amused by something {\textcolor{blue}{[BLANK]}} off-screen.} & 
1. \textcolor{gray}{A bald headed man in business casual attire \highlightres{isn't} amused by something \highlightres{EMPTY} off-screen.} \newline \newline & 0.789 & 0.756 & 278.892 \vspace{0.6cm} \\ \hline
\end{tabular}} }
\caption{Comparison of model outputs across different inputs with associated metrics. This table contrasts the generated sentences with their original counterparts, showcasing variations in grammaticality (Gramm.), fluency (Flu.), and perplexity (PPL). It includes examples of degenerate text and outputs with empty tokens, highlighting issues such as repetition, low coherence, and incomplete responses.}
\label{tab:model_corrupted}
\end{table}

\begin{table}[h]
\centering
\scalebox{0.8}{
\begin{tabular}{@{}l|p{0.98\textwidth}@{}}
\toprule
\multicolumn{2}{c}{\textbf{\methodName Filtering Examples}} \\
\midrule

\textbf{Original:} & They remained loyal to their cause despite the challenges. \\
\hline
\textbf{Generated:} & 
\begin{tabular}[c]{@{}l@{}}
    Not remained loyal to their cause despite the challenges.\\
    None of them remained loyal to their cause despite the challenges. \\
    Not as long as they remained loyal to their cause despite the challenges. \\
    \sout{They remained loyal to their cause despite the challenges..} \\
    \sout{They remained loyal to their cause despite \textcolor{red}{lack of} challenges..} \\
    \sout{They remained loyal to their cause despite \textcolor{red}{lack of} adversity..} \\
    \sout{They remained loyal to their cause despite \textcolor{red}{lack of} adversity..} \\
    \sout{They remained loyal to their cause despite their struggles..} \\
    They remained indifferent to their cause despite the challenges. \\
    They remained not loyal to their cause despite the challenges. \\
    They remained dispirited to their cause despite the challenges. \\
    They did not remain loyal to their cause despite the challenges. \\
    They didn't remain loyal to their cause despite the challenges. \\
    They weren't loyal to their cause despite the challenges.\\
    \sout{Not remained loyal to their cause despite the challenges.} \\
    They never remained loyal to their cause despite the challenges. \\
    \sout{None of them remained loyal to their cause despite the challenges.}
\end{tabular} \\
\hline
\textbf{Filtered:} & 
\begin{tabular}[c]{@{}l@{}}
    \textcolor{gray}{\highlightres{None of them} remained loyal to their cause despite the challenges} \\
    \textcolor{gray}{They remained \highlightres{indifferent} to their cause despite the challenges} \\
    \textcolor{gray}{They \highlightres{didn't} remain loyal to their cause despite the challenges} \\
    \textcolor{gray}{They \highlightres{never} remained loyal to their cause despite the challenges} \\
    \textcolor{gray}{They \highlightres{weren't} loyal to their cause despite the challenges} \\
    \textcolor{gray}{\highlightres{Not} remained loyal to their cause despite the challenges} \\
    \textcolor{gray}{They remained \highlightres{dispirited} to their cause despite the challenges} \\
    \textcolor{gray}{\highlightres{Not as long as} they remained loyal to their cause despite the challenges} \\
    \textcolor{gray}{They \highlightres{did not} remain loyal to their cause despite the challenges} \\
    \textcolor{gray}{They remained \highlightres{not} loyal to their cause despite the challenges}
\end{tabular} \\

\bottomrule
\end{tabular}}
\caption{This table compares the original sentence with perturbations generated by \methodName and the final filtered versions. The \textbf{Original} text serves as the reference. The \textbf{Generated} section displays different sentences produced by the model, with red text (\textcolor{red}{text}) indicating negation cues that were not detected by the negation detector. The \textbf{Filtered} section shows sentences selected based on criteria such as the elimination of repeated sentences, removal of sentences without negations, and filtering based on a Levenshtein distance threshold. Key terms in the filtered sentences are highlighted as negation cues in purple (\highlightres{purple}). These examples showcase the effectiveness of the filtering criteria and highlight discrepancies in negation detection, particularly where NegBERT fails to correctly detect affixes and multi-word negation cues.}
\label{tab:filter_design}
\end{table}

 \paragraph{On Perplexity Being a Misleading Metric.} In Table \ref{tab:evaluation}, we show the impact of word choices on perplexity (PPL), fluency, and grammaticality in text generation. For instance, the sentence "\textit{Her sweater is uncomfortable and pretty}" scores high in fluency and grammaticality but has a notably high PPL. The increased perplexity suggests that the word "\texttt{\color{CadetBlue}uncomfortable}", despite its grammatical correctness and naturalness, is less predictable for the model. This may be due to the less frequent occurrence of affixal negation -- such as "\texttt{\color{CadetBlue}un-}" -- in GPT-2's training data, making such constructions more challenging, especially when paired with a positive attribute like "\texttt{\color{CadetBlue}pretty}". When the sentence is rephrased to "\textit{Her sweater is not comfortable and pretty}", the PPL drops significantly, indicating a more predictable structure for the model. However, this rephrasing results in slightly lower fluency and grammaticality scores.

The sentence "\textit{He doesn’t offer a rational explanation for his decision}" scores high in both fluency and grammaticality with a very low perplexity, demonstrating a case where low perplexity correlates well with high-quality metrics. In contrast, the sentence "\textit{She spent the day wearing \textcolor{red}{nothing sweater}}" exhibits high perplexity but maintains an unusually high fluency score, despite being nonsensical. This discrepancy indicates that perplexity and fluency scores may not always align with human judgment. Additionally, the phrase \textit{\textcolor{red}{"didn't slept"}} has a higher grammaticality score than that of \textit{\textcolor{red}{"none of the kids"}}, highlighting that these metrics do not always capture grammatical nuances accurately.

These examples indicate that PPL and and quality scores can be useful as a general measure of a model's predictive capabilities, but they should not be used in isolation to assess the naturalness and coherence of the generated data.

\paragraph{Degenerate Cases Analysis for NegVerse.}
As noted earlier in the limitations section, despite \methodName's strong performance in preserving syntactic structure and offering a greater variety of negation types, it occasionally produces degenerate outputs. These issues are particularly evident with blank placements at the end of sentences, sometimes leading to grammatically correct but contextually meaningless results. We provide a number of degenerate output examples in Table~\ref{tab:model_corrupted}. 

The first example of the table illustrates a case where the model generates a sequence of special characters, such as \textcolor{red}{\texttt{|> [|> [|> [|> [|> [|> [|> [|> [|> [|> [|>}}, in response to the prompt. This output is marked by a low grammaticality score of 0.769 and a fluency score of 0.760, indicating deficiencies in both grammatical correctness and fluency. Despite these low scores, the perplexity is notably low, suggesting that the model finds this sequence statistically probable, although the output remains largely nonsensical.

In contrast, for the same input, the well-formed output "\textit{He offers a rational explanation for his decision.}" achieves high grammatical and fluency scores, but exhibits a higher perplexity compared to degenerations. It is noteworthy that degenerate outputs occur in instances where "{\textcolor{blue}{[BLANK]}}" appears at the end of the sentence. Removing the period, as seen in the case with "{\textcolor{blue}{[BLANK]}}", allows the model to generate a correct output, indicating that the presence of the period may contribute to issues in the generation process.

The second example in Table~\ref{tab:model_corrupted} demonstrates an issue with repeated text. In this case, the model generates a response that includes repeated and incoherent phrases, such as \textit{\textcolor{red}{" not a young woman.......... not a young woman.............. not a young woman........"}}. This repetitive output is truncated in the table for visibility but illustrates a broader problem with the model's generation process. The repetition contributes to a low perplexity but results in a lack of coherence and meaningful content.

In the output example featuring the empty token, "{\textcolor{blue}{[EMPTY]}}" is used as a placeholder to represent missing or unspecified content. This indicates that the model was unable to generate a specific word or phrase, leading to a vague or incomplete response. The use of the empty token highlights a limitation in the model’s ability to produce coherent text in certain contexts. Additionally, not every position in a sentence is suitable for introducing a negation, which further contributes to the model’s challenges in generating appropriate and contextually accurate content.

\paragraph{Examples of  \methodName Application.}
In Table~\ref{tab:filter_design}, we provide an example of the filtered results that were selected from sentences generated by the model using Algorithm \ref{alg:select_sentences}. Recall that the proposed filtering mechanism uniformly samples from sentences containing effective negations close to the original affirmative sentence. As shown in the provided example, \methodName sometimes misses certain negation cues, such as "\texttt{\color{CadetBlue}lack of}" in specific contexts. Nevertheless, the model successfully identified other types of negation, such as "\texttt{\color{CadetBlue}indifferent}" and "\texttt{\color{CadetBlue}dispirited}" which, although not direct affixal negations of "\texttt{\color{CadetBlue}loyal}" are still relevant for expressing negation. This behaviour may stem from the model's limited training data on affixal negations and insufficient exposure to diverse contexts. Despite these limitations, the model effectively detects most non-verbal negations, such as "\texttt{\color{CadetBlue}never}" and "\texttt{\color{CadetBlue}none of}" as well as verbal forms like "\texttt{\color{CadetBlue}didn't}" and "\texttt{\color{CadetBlue}weren't}".
\begin{table}[h]
\centering { 
\begin{tabular}{@{}l|p{0.8\textwidth}@{}}
\toprule
\multicolumn{2}{c}{\textbf{\methodName Generations with {\textcolor{blue}{[BLANK]}}}} \\
\midrule

\textbf{Original:} & She is always happy to lend a helping hand to her friends. \\
\textbf{Generated:} & ['She is never happy to lend a helping hand to her friends.', 'She is not always happy to lend a helping hand to her friends.', 'She is not happy to lend a helping hand to her friends.'] \\
\textbf{\textbf{Filtered:}} & \textcolor{gray}{['She is \highlightres{never} happy to lend a helping hand to her friends', 'She is \highlightres{not} happy to lend a helping hand to her friends', 'She is\highlightres{not} always happy to lend a helping hand to her friends']} \\

\midrule

\textbf{Original:} & The design makes the new car highly desirable. \\
\textbf{Generated:} & ['The design makes the new car highly undesirable.', 'The design makes the new car highly desirable.', 'The design makes the new car highly undesirable.', 'The design makes the new car highly un desirable.'] \\
\textbf{\textbf{Filtered:}} & \textcolor{gray}{['The design makes the new car highly \highlightres{un desirable}', 'The design makes the new car highly \highlightres{undesirable}']} \\

\midrule

\textbf{Original:} & They remained loyal to their cause despite the challenges. \\
\textbf{Generated:} & ['They remained loyal to their cause despite the challenges.', 'They remain loyal to their cause despite the challenges.', 'They did not remain loyal to their cause despite the challenges.'] \\
\textbf{\textbf{Filtered:}} & \textcolor{gray}{['They \highlightres{did not} remain loyal to their cause despite the challenges']} \\

\midrule

\textbf{Original:} & Technology allows us to connect with people across the globe instantly. \\
\textbf{Generated:} & ['Technology allows us to connect with people across the globe instantly.', 'Technology allows us to connect with people across the world.', 'Technology allows us to connect with people across the globe.'] \\
\textbf{\textbf{Filtered:}} & \textcolor{gray}{[]} \\

\midrule

\textbf{Original:} & The cat napped peacefully. \\
\textbf{Generated:} & ['The cat napped peacefully.', 'The cat napped peacefully..', 'The cat did not nap peacefully.'] \\
\textbf{\textbf{Filtered:}} & \textcolor{gray}{['The cat \highlightres{did not} nap peacefully']} \\

\bottomrule
\end{tabular}}
\caption{Generated and filtered outputs for affirmative sentences where the entire sentence is masked. The model tends to negate only a single part of the sentence rather than introducing diverse perturbations. Additionally, as the length of the sequence increases, the performance of the model in negating the sentence deteriorates, resulting in cases where the model simply repeats the original sentence, leading to no new or meaningful output. The negation cue produced by \methodName is \highlightres{highlighted}.}

\label{tab:filtered_blank}
\end{table}

\paragraph{\methodName Generation with {\textcolor{blue}{[BLANK]}} for Complete Sentence Masking.} In Table~\ref{tab:filtered_blank} we show examples where the model generates negated sentences without explicit guidance on blank placement. The results show that the model can produce various negations for simpler sentences effectively. However, the model's performance becomes inconsistent with more complex sentences, leading to issues such as repetition or awkward phrasing. This variability indicates that while the model handles basic negations well, its ability to consistently apply negation across different sentence structures without precise blank placement can be limited.

\subsection{Evaluation Metrics}
In this section, we provide further details on the evaluation metrics used in the main manuscript to assess the performance of both the proposed approach and the baseline methods.
We consider metrics to examine various aspects of negated text generation, including closeness, fluency, and diversity. 

\paragraph{(Average) Levenshtein Distance (NLD):} This metric measures the average minimum number of edits needed to transform one tokenized sentence into another. The formal definition is provided below:
 
\[
\text{NLD} = \frac{1}{N} \sum_{i=1}^N \frac{d(x_i,\hat{{x}}_{\textsf{gen},i} )}{\max(|x_i|, |\hat{{x}}_{\textsf{gen},i}|)}
\]

where $x_i$ denotes the reference sentence, $\hat{{x}}_{\textsf{gen},i}$ is the generated negated sentence from the model, and $n$ is the total number of sentence pairs. This metric has been widely used in various studies to evaluate the similarity between sentence pairs, particularly in counterfactual evaluations.\cite{nguyen2024llmsgeneratingevaluatingcounterfactuals,ross2021explainingnlpmodelsminimal,treviso2023crestjointframeworkrationalization}.

\paragraph{Self-BLEU Score:} This metric evaluates the diversity within a set of generated texts by measuring their similarity to each other, as opposed to traditional BLEU, which compares generated texts to reference texts \cite{zhu2018texygenbenchmarkingplatformtext}. The Self-BLEU score is calculated as:
\[
\text{Self-BLEU} = \frac{1}{m} \sum_{i=1}^{m} \text{BLEU}\left(\hat{x}_{\textsf{gen},i}, \hat{\mathcal{X}}_{\textsf{gen}} \setminus \{\hat{x}_{\textsf{gen},i}\}\right) 
\]

where $m$ is the total number of generated sentences, $\hat{x}_{\textsf{gen},i} $ is the $i$-th generated sentence, and $\hat{\mathcal{X}}_{\textsf{gen}} \setminus \{\hat{x}_{\textsf{gen},i} \}$ represents the set of all generated sentences except $\hat{x}_{\textsf{gen},i} $. A lower Self-BLEU score indicates higher diversity, while a higher score suggests more similarity among outputs.
 
\textbf{Perplexity:} This metric evaluates how well a language model predicts a sequence of tokens, with lower perplexity indicating better fluency. It has been widely used for fluency assessment in text generation models like GPT-$2$ \cite{nguyen2024llmsgeneratingevaluatingcounterfactuals, treviso2023crestjointframeworkrationalization}. For a negated sentence $\hat{x} = (\hat{z}_1, \hat{z}_2, \ldots, \hat{z}_n)$, where $n$ is the sentence length, the perplexity $\text{PPL}(\hat{x})$ is given by:

\[
\text{PPL}(\hat{x}) = \exp \left( -\frac{1}{n} \sum_{i=1}^{n} \log p_\theta(\hat{z}_i \mid \hat{z}_{<i}) \right)
\]

where $\log p_\theta(\hat{z}_i \mid \hat{z}_{<i})$ is the log probability of token $\hat{z}_i$ given the preceding tokens $\hat{z}_{<i}$.

\end{document}